\documentclass[11pt]{article}
\usepackage[T1]{fontenc}
\usepackage{tikz}
\usepackage{amsmath}
\usepackage{booktabs}
\usepackage{algpseudocode}
\usepackage[ruled]{algorithm2e}
\usepackage{amsmath}
\usepackage{mathtools}
\usepackage{subcaption}
\usepackage{fullpage}
\usepackage{amssymb}
\usepackage{esvect}
\newcommand{\scpr}[1]{\left\langle #1 \right\rangle}
\newcommand{\bb}[1]{\mathbb{#1}}

\newcommand{\comp}{\ensuremath{K}}
\renewcommand{\d}{\mathrm{d}}
\mathtoolsset{showonlyrefs}

\newcommand*{\e}{\ensuremath{\mathrm{e}}}

\usepackage{svg}

\renewcommand*{\rho}{\varrho}

\newcommand*{\RR}{\ensuremath{\mathbb{R}}}

\title{Sampling from Boltzmann densities with physics informed low-rank formats}

\author{Paul Hagemann$^1$, Janina Schütte$^2$, David Sommer$^2$, Martin Eigel$^2$, Gabriele Steidl$^1$ \\
$^1$ Technische Universit\"at Berlin,
  $^2$ Weierstraß Insitute of Applied Analyis and Stochastics\\
  Correspondence to: \texttt{hagemann@math.tu-berlin.de, $\{$schuette,sommer$\}$@wias-berlin.de}}

\begin{document}

\maketitle

\begin{abstract}
    Our method proposes the efficient generation of samples from an unnormalized Boltzmann density by solving the underlying continuity equation in the low-rank tensor train (TT) format. It is based on the annealing path commonly used in MCMC literature, which is given by the linear interpolation in the space of energies. Inspired by Sequential Monte Carlo, we alternate between deterministic time steps from the TT representation of the flow field and stochastic steps, which include Langevin and resampling steps. These adjust the relative weights of the different modes of the target distribution and anneal to the correct path distribution. We showcase the efficiency of our method on multiple numerical examples.
\end{abstract}

\section{Introduction}
Sampling from Boltzmann densities with unknown normalization constant is an important and timely research problem~\cite{Albergo_2019,Del_Debbio_2021,noe_boltzmann}. The main challenge is that no samples are given. It hence is difficult to explore the probability space where the mass is concentrated. In the context of MCMC, this leads to the so-called energy barriers \cite{mcmc_barrier}, which means that it takes chains a long time to jump from one mode of the distribution to another with uninformed proposals. In the context of (conditional) normalizing flows, the training of sampling problems is usually carried out using the reverse Kullback-Leibler (KL) divergence~\cite{noe_boltzmann}, which leads to unwanted mode-collapse. This in principle can be fixed by using MCMC steps to either favor exploration via path learning ideas~\cite{arbel2021annealed,HHS2022,wu2020stochastic} or by considering a mode-covering objective~\cite{midgley2023flowannealedimportancesampling}.

Recently, the path-based learning objective literature has not only been explored via variational objectives, which essentially rely on reformulations of the KL divergence, but also by learning the distributions via the corresponding PDEs.
This can for instance be the continuity equation~\cite{balint} or the Hamilton-Jacobi-Bellman equation~\cite{berner2024an}. 
Conceptionally, it is often done in a way similar to physics-informed neural networks (PINN)~\cite{RAISSI2019686}, which is known to suffer from stability and optimization issues \cite{chuang2022experience} as well as slow training.

In this work, we address these shortcomings. Concretely, we learn the velocity field of the continuity equation inspired by~\cite{balint}, which was refined in~\cite{sun2024dynamicalmeasuretransportneural}.
To overcome stability issues and speed up training, we explore the tensor train (TT) framework~\cite{holtz2012manifolds,O11,oseledets2013constructive}, which has shown great success in solving PDEs~\cite{EigelNeumannSchneiderWolf+2019+39+53,dolgov2021tensor,richter2021solving,vemuri2024functionaltensordecompositionsphysicsinformed}, and which has recently been used for sampling with diffusion paths \cite{sommer2024TT}. Here, we follow a classical approach often used in Sequential Monte Carlo methods \cite{del2006sequential,doucet2000sequential,neal2001annealed} and reformulate the (discrete) continuity equation by expansion into a product basis. This reduces training to solving linear systems of equations. To facilitate expressivity, we enrich the flow with stochastic steps combining ideas from~\cite{arbel2021annealed,doucet2000sequential,wu2020stochastic}, in particular resampling and Langevin steps. In contrast to \cite{arbel2021annealed}, we replace their normalizing flow layers with TT constructions with associated functional basis.

Our contributions can be summarized as follows:
\begin{itemize}
    \item We introduce a functional tensor train (FTT) format for solving the continuity equation for the annealing path. In particular, we derive formulas for the action of the continuity equation on vector fields represented by FTTs with arbitrary basis functions.
    \item Using the action of the continuity equation on FTT, we propose a numerical optimization scheme involving empirical risk minimization with the alternating linear scheme (ALS) \cite{holtz2012alternating}. Our approximation is naturally rank-adaptive due to an exponential moving average of successive TT approximations. 
    \item We apply an $H^2$ (Sobolev) orthonormal Fourier basis opposed to the commonly used polynomials within the TT format~\cite{dolgov2021tensor,eigel2023dynamical,richter2021solving}. 
    \item We add stochastic steps, namely Langevin postprocessing and resampling, to improve expressivity. 
    \item We illustrate the efficiency of the format on two-dimensional Gaussian mixture problems and the ``many well problem'' introduced in~\cite{midgley2023flowannealedimportancesampling}.
    \end{itemize}

\section{Logarithmic Continuity Equation}
We want to sample from Boltzmann densities depending on functions $f:[0,1]\times \RR ^d \to \RR$ satisfying mild conditions such that for $f_t(\cdot)\coloneqq f(t,\cdot)$ 
\begin{equation} \label{boltz}
    p_t \coloneqq \frac{e^{-f_t}}{Z_t}, \quad Z_t \coloneqq \int_{\RR^d} e^{-f_t} \, \d x.
\end{equation}
We follow the considerations in \cite{chemseddine2024neuralsamplingboltzmanndensities,balint} and refer to \cite{chemseddine2024neuralsamplingboltzmanndensities} for a mathematically rigorous treatment.
Starting with an easy-to-sample latent distribution $p_0$, the goal is to sample from a target distribution $p_1$ knowing only $f_1$.
In this paper, we follow the annealing path
\begin{equation} \label{inter}
    f_t = t f_1 + (1-t) f_0, \quad t \in [0,1],
\end{equation}
which is commonly used in the MCMC literature \cite{arbel2021annealed,doucet2000sequential,neal2001annealed,wu2020stochastic}.
Now the task is to find a measurable function $v: [0,1] \times \RR^d \to \RR^d$ 
% with $\|v_t\|_{L^2(\RR^d,p_t)} \in L^1((0,1])$ 
such that $(p_t,v_t)$ satisfies the continuity equation
\begin{align} \label{eq:ce}
    \partial_t p_t + \nabla \cdot (p_t v_t) = 0.
\end{align}
%Then the curve $p_t:[0,1] \to \mathcal P(\RR^d)$ is an absolutely continuous curve in the Wasserstein geometry \cite{ambrosio2008gradient}.
For the Boltzmann density, the continuity equation becomes \cite{balint}
\begin{align*}
    \big(\partial_t f_t + \langle \nabla f_t, v_t\rangle - \nabla \cdot v_t + C_t \big) p_t= 0 
    \quad \text{with} \quad
    C_t := \partial_t(\log Z_t).
\end{align*}
With the linear interpolation \eqref{inter} we obtain
\begin{align} \label{eq:final}
    f_1 - f_0 + \langle \nabla f_t v_t\rangle - \nabla \cdot v_t + C_t = 0 \quad \text{a.e. } t \in [0,1].
\end{align}
Indeed, it was shown under mild assumptions on $p_0$ and $p_1$ that appropriate Borel measurable vector fields $v_t$ exist, see \cite{chemseddine2024neuralsamplingboltzmanndensities} and the references therein.
Fortunately, absolutely continuous curves $p_t$ with corresponding vector fields $v_t$ in \eqref{eq:ce} can be described by the solution $\phi:[0,1] \times \RR^d \to \RR^d$ of an ODE 
\begin{equation}\label{eq:dynamics}
    \partial_t \phi(t,x) = v_t(\phi(t,x)), \quad \phi(0,x) = x
\end{equation}
and $\phi(t, \cdot)_{\#} p_0 = p_t$, see \cite[Theorem 8.1.8]{ambrosio2008gradient}.
Therefore, once the velocity field satisfying \eqref{eq:final} is known,
one can sample from $p_1$ by solving the above ODE with classical numerical solvers.
In this paper, we propose to approximate $v_t$ by a low-rank TT representation $v^\theta_t$ that minimizes by \eqref{eq:final} the least squares problem
$$
    \mathcal J_t(\theta) \coloneqq |f_1 - f_0 + \langle \nabla f_t ,v_t^\theta\rangle - \nabla \cdot v_t^\theta + C_t|^2.
$$

%-----------------------------------------------------
\section{Low-rank tensor train decomposition}

We propose to approximate the vector field $v_t$ with a tensor product ansatz.
Choosing a basis $\{ \varphi_{i}: \mathbb{R}\to\mathbb{R}: i \in [n]\}$, $[n]:=\{1,\ldots,n\}$, of a space of univariate functions, $v_t^\theta: \RR^d \to \RR^d$ is then represented in a product basis expansion.
Collecting the basis coefficients in an array yields a high-dimensional tensor $T\in\bb{R}^{d\times n\times\dots \times n}$ with $dn^d$ entries.
Hence, the storage complexity quickly becomes prohibitive in higher dimensions. To alleviate this, an approximation is directly learned in a lower dimensional tensor manifold. There are different low-rank formats used in the literature, such as the canonic-polyadic (CP), the hierarchical (HT) or TT format. For a comprehensive overview, we refer to \cite{Nouy_2017}.
%The TT format has been known in the physics community as \textit{matrix product states}. 
The TT format is particularly useful for approximation tasks: contrary to the CP format, the set of tensors with so-called ``TT rank'' smaller than a fixed upper bound is closed.
Moreover, there exists a gradient-free alternating optimization procedure known as \textit{alternating linear scheme} (ALS), which is used in this work.

For a formal derivation of the TT format, we note that each entry of a tensor $T\in\bb{R}^{d\times n\times\dots \times n}$ can be represented by a multiplication of matrices. For a multi-index $(i_1,\dots,i_d) \in [n]\times \dots \times [n]$, we write
\begin{equation}\label{ttc}
    T_{:,i_1,\dots,i_d} = \comp^{(1)}_{i_1} \cdot \ldots \cdot \comp^{(d)}_{i_d} \in\mathbb{R}^d,
\end{equation}
where $\comp^{(1)}_{i_1}\in\bb{R}^{d\times r_1}$, $\comp^{(d)}_{i_d}\in \bb{R}^{r_{d-1}\times 1}$ and $\comp_{i_k}^{(k)}\in \bb{R}^{r_{k-1}\times r_k}$ for $k=2,\dots, d-1$ are matrices.
The smallest values $r_1,\dots,r_{d-1}\in\bb{N}$ such that a representation \eqref{ttc} exists are called the \emph{TT-ranks}. 
The three dimensional tensors $\comp^{(k)}\in\bb{R}^{r_{k-1} \times n \times r_k}$ %$\comp^{(k)}\in\bb{R}^{r_{k-1} \times {\color{red} d} \times r_k}$
containing the matrices $\comp^{(k)}_{i_k}$ for $k\in [d]$ and $i_k\in[n]$ are called the \emph{cores} (or \emph{component tensors}) of the TT.
Using a graphical notation,
we can write the tensor as a contraction of the TT cores along the rank dimensions
\begin{center}
    \begin{tikzpicture}[scale=0.2,yscale=1,xscale=1,thick,
    mynode/.style={rectangle, rounded corners=0.5ex, draw=red!60, fill=red!5, very thick, minimum size=5mm},
    mynodex/.style={rectangle, rounded corners=1ex, draw=green!60, fill=green!5, very thick, minimum size=5mm}
    ]
     
    \node[mynode] at (47.0,0) {$T$};
    \draw (45.5,0)--(44.00,0);
    \draw (48.5,0)--(50.00,0);
    \draw (47.0,-1.5)--(47.0,-3);
    \draw (47.0,+1.5)--(47.0,+3);
    \draw (47.0,+1.5)--(47.0,+3);
    \node[right] at (47.25,2.25) {$\ddots$};
    \node[right] at (44.0,-3.0) {$d$};
    \node[right] at (42.0,1.0) {$n$};
    \node[right] at (44.0,3.0) {$n$};
    \node[right] at (48.75,-1.0) {$n$};
    
    \node at (61.0,0) {$\longrightarrow$};
    \end{tikzpicture}
    \qquad\begin{tikzpicture}[scale=0.2,yscale=1,xscale=1,thick,
    mynode/.style={rectangle, rounded corners=1ex, draw=blue!60, fill=blue!5, very thick, minimum size=5mm},
    mynodex/.style={rectangle, rounded corners=1ex, draw=green!60, fill=green!5, very thick, minimum size=5mm}
    ]
    
    % \node[left] at (-2,0) {$T = $};

    \draw (-2,0)--(1,0);
    \node[above] at (-1.5,0) {$d$};
    \node[mynode] at (2,0) {$\comp^{(1)}$};
    \draw (4.5,0)--(8,0);
    \node[above] at (6,0) {$r_1$};
    \draw (2,-1.5)--(2,-3);
    \node[left] at (2,-3) {$n$};
    
        \node[mynode] at (10,0) {$\comp^{(2)}$};
    \draw (12.5,0)--(14.5,0);
    \node[above] at (14.5,0) {$r_2$};
    \draw (10,-1.5)--(10,-3);
    \node[left] at (10,-3) {$n$};
    
    \node at (17.5,0) {\dots};
    
    \node[mynode] at (25.5,0) {$\comp^{(d)}$};
    \draw (22.75,0)--(20.75,0);
    \node[above] at (20.5,0) {$r_{d-1}$};
    \draw (25.5,-1.5)--(25.5,-3);
    \node[right] at (25.5,-3) {$n$};
    
    \node at (29.5,0) {.};
    \end{tikzpicture}
\end{center}

The basis expansion with coefficient tensor $T$ can be rewritten based on the TT representation for any $x = (x_1,\dots,x_d)^\intercal \in\mathbb{R}^d$ as follows:
\begin{small}
\begin{equation*}
    v(x) = \sum_{i_1=1}^{n_1} \cdots \sum_{i_d=1}^{n_d} \textcolor{red}{T_{:,i_1,\ldots,i_d}} \prod_{k=1}^d \varphi_{i_k}(x_k) \quad \longrightarrow \quad  v(x) = \sum_{i_1=1}^{n_1} \cdots \sum_{i_d=1}^{n_d} \textcolor{blue}{ \prod_{k=1}^d \comp^{(k)}_{i_k}} \varphi_{i_k}(x_k).
\end{equation*}
\end{small}
Imposing a low-rank structure on the coefficient tensor $T$ results in a manifold of functions with low-rank tensors of rank at most $\mathbf r = (r_1,\ldots,r_{N-1})$.
In practice this can be understood as neglecting spurious features by an implicit feature selection.
Analytical bounds, which quantify this behavior rigorously are given in~\cite{O11}. %,OT10}. 
Explicit constructions of some notable functions in TT format can be found in \cite{oseledets2013constructive}. The combination of a TT with an associated set of basis functions defines an FTT.

\subsection{Vector field learning with the alternating linear scheme}\label{sec:als}

We approximate the vector field $v_t$ at fixed times $t$ as an FTT $v_t^{\theta}$ of order $d$ and with $d$-dimensional output. The parameter $\theta$ is short-hand notation for the collection of all TT components $\comp_t^{(1)},\ldots,\comp_t^{(d)}$ defining $v_t^{\theta}$. The FTT is the contraction of the TT components with the basis functions $\varPhi(x_k) \coloneqq (\varphi_{1}(x_k),\ldots,\varphi_{n}(x_k))^T$ for each $k\in [d]$, yielding a $d$-dimensional output,
\begin{center}
    \begin{tikzpicture}[scale=0.2,yscale=1,xscale=1,thick,
    mynode/.style={rectangle, rounded corners=1ex, draw=blue!60, fill=blue!5, very thick, minimum size=5mm},
    mynodex/.style={rectangle, rounded corners=1ex, draw=green!60, fill=green!5, very thick, minimum size=5mm}
    ]
    
    \node[left] at (-2.5,0) {$v_t^\theta(x) = $};
    
    \draw (-2,0)--(1,0);
    \node[above] at (-1.5,0) {$d$};
    \draw (4.5,0)--(8,0);
    \node[above] at (6,0) {$r_1$};
    \draw (2,-1.5)--(2,-4);
    \node[left] at (2,-3) {$n$};
    \node[mynode] at (2,0) {$\comp_t^{(1)}$};
    
    \draw (12.5,0)--(14.5,0);
    \node[above] at (14.5,0) {$r_2$};
    \draw (10,-1.5)--(10,-4);
    \node[left] at (10,-3) {$n$};
    \node[mynode] at (10,0) {$\comp_t^{(2)}$};
        
    \node at (17.5,0) {\dots};
    
    \draw (22.75,0)--(20.75,0);
    \node[above] at (20.5,0) {$r_{d-1}$};
    \draw (25.5,-1.5)--(25.5,-4);
    \node[right] at (25.5,-3) {$n$};
    \node[mynode] at (25.5,0) {$\comp_t^{(d)}$};

    \node[mynodex] at (25.5,-6) {$\varPhi(x_d)$};
    \node[mynodex] at (10,-6) {$\varPhi(x_2)$};
    \node[mynodex] at (2,-6) {$\varPhi(x_1)$};
    
    \node at (29.5,0) {.};
    \end{tikzpicture}
\end{center}
For the training process, the empirical $L^2$ error in \eqref{eq:final}
on points $\{x^{(\ell)}\}_{\ell=1}^N$ with
\begin{equation}\label{eq:emp_error}
   J_{t,N}(\theta) =  \sum_{\ell=1}^N \left| \partial_t f_t(x^{(\ell)}) + \scpr{\nabla f_t(x^{(\ell)}),v^{\theta}_t(x^{(\ell)})}  - \nabla \cdot v^{\theta}_t(x^{(\ell)}) + C_t \right|^2
\end{equation}
is minimized. In the alternating linear scheme \cite{holtz2012alternating}, the mappings $\comp_t^{(i)} \mapsto J_{t,N}(\theta)$ are iteratively minimized over $\comp_t^{(i)}$, while keeping the other components %\\ $\comp_t^{(1)},\ldots \comp_t^{(i-1)},\comp_t^{(i+1)},\ldots,\comp_t^{(d)}$ 
fixed. Once $\comp_t^{(i)}$ is optimized, the next component $\comp_t^{(i+1)}$ is optimized and so on, sweeping forward and backward through the chain of TT components until a (local) minimum is reached. 

%In this way, minimizing the empirical error functional \eqref{eq:emp_error} over $X_t^{(i)}$ while keeping the other components of $v_t^{\theta}$ fixed can be expressed as
By \eqref{eq:emp_error}, there exist linear operators 
$\mathcal{L}_i\colon \mathbb{R}^{r_{i-1}\times n \times r_i}%\mathcal{L}_i\colon \mathbb{R}^{r_{i-1}\times {\color{red} n} \times r_i}
\to \mathbb{R}^N$ such that the minimization over one core can be written as
\begin{equation}\label{eq:quad_program}
    \min_{K_t^{(i)}} J_{t,N}(v_t^{\theta}) = \min_{K_t^{(i)}}\left\| \mathcal{L}_i(\comp_t^{(i)})-y \right\|^2,
\end{equation}
where $y\in\mathbb{R}^N$ with $y_\ell = -\partial_t f_t (x^{(\ell)})-C_t$ and the entries of $\mathcal{L}_i(\comp_t^{(i)})$ %$ = (\mathcal{L}_{i,x^{(1)}}(X_t^{(i)}),\ldots,\mathcal{L}_{i,x^{(N)}}(X_t^{(i)}))^T$ % , $y = (\partial_t f_t (x^{(1)})-C_t,\ldots,\partial_t f_t (x^{(N)})-C_t)^T\in\mathbb{R}^N$ 
%and
% the linear operators $\mathcal{L}_{i,x}$ 
are defined for $\ell=1,\dots,N$ in graphical notation by 
\begin{center}
\scalebox{0.7}{
    \begin{tikzpicture}[scale=0.2,yscale=1,xscale=1,thick,
    mynode/.style={rectangle, rounded corners=1ex, draw=blue!60, fill=blue!5, very thick, minimum size=5mm},
    mynodex/.style={rectangle, rounded corners=1ex, draw=green!60, fill=green!5, very thick, minimum size=5mm}
    ]
    
    \node[left] at (-8,-11) {$\mathcal{L}_{i}(K_t^{(i)})_\ell=$};
    
    \node[left] at (-2.5,-11) {$\sum_{k=1}^d $};
    
    % \draw (-2,-11)--(1,-11);
    % \node[above] at (-1.5,-11) {$k$};
    \node[mynode] at (2,-11) {$\comp^{(1)}_t[k]$};
    \draw (5.5,-11)--(7,-11);
    \node[above] (s) at (6.75,-11) {$r_1$};
    \draw (2,-12.5)--(2,-15);
    \node[left] at (2,-14) {$n_1$};
    
    \node at (9,-11) {\dots};

    \draw (11,-11)--(28,-11);
    \node[mynode] (s) at (19.5,-11) {$\comp_t^{(k)}$};
    \node[above] at (11,-11) {$r_{k-1}$};
    \node[above] at (28,-11) {$r_{k}$};
    \draw (19.5,-12.5)--(19.5,-15);
    \node[right] at (19.5,-14) {$n$};

    \draw[dashed] (-2.5,-8) rectangle (31.33,-23.5);
    \draw[dashed] (46.5,-8) rectangle (60.33,-23.5);
    \draw[dashed] (32.5,-15.5) rectangle (45.33,-23.5);

    \node at (30,-11) {\dots};
    \node at (33,-11) {\dots};

    \draw (34.5,-11)--(39,-11);
\node[mynode] (s) at (39.5,-11) {$\comp_t^{(i)}$};
\node[above] at (35.5,-11) {$r_{i-1}$};
\draw (39.5,-12.75)--(39.5,-14.25);
\node at (39.5,-15) {$\vdots$};
\draw (39.5,-17)--(39.5,-19);
\node[right] at (39.5,-14) {$n$};

\node[mynodex] at (39.5,-20) {$\varPhi\left(x_i^{(\ell)}\right)$};
\node[mynodex] at (19.5,-17) {$\left(\partial_k f_t\left(x^{(\ell)}\right) - \partial_k\right) \varPhi\left(x_k^{(\ell)}\right)$};
\node[mynodex] at (2,-17) {$\varPhi\left(x_1^{(\ell)}\right)$};

\node[above] at (43,-11) {$r_{i}$};
\draw (41.75,-11)--(43,-11);
\node at (45,-11) {\dots};
\draw (49.75,-11)--(52.25,-11);

\node at (48,-11) {\dots};
\node[mynode] (s) at (54.5,-11) {$\comp_t^{(d)}$};
\node[above] at (50.5,-11) {$r_{d-1}$};
\draw (54.5,-12.5)--(54.5,-15);
\node[right] at (54.5,-14) {$n$};

\node[mynodex] at (54.5,-17) {$\varPhi\left(x_d^{(\ell)}\right)$};

\node at (62,-11) {.};
\end{tikzpicture}
}
\end{center}
Here, $\comp_t^{(1)}[k]$ denotes the $\mathbb{R}^{1\times r_1}$-vector %$\mathbb{R}^{{\color{red} n}\times r_1}$-matrix 
defined by $(\comp_t^{(1)}[k])_{i,j} \coloneqq {\comp_t^{(1)}}_{k,i,j}$.
Due to the linearity of $\mathcal{L}_i$, \eqref{eq:quad_program} is a simple quadratic program, which can be solved by standard techniques. The ALS solves this quadratic program iteratively for all components until convergence. The convergence of ALS to local minima is investigated in \cite{rohwedder2013local}. 

\section{Methodology}

The vector field $v_t = v_t^{\theta}$ is learned on a fixed time grid $t_0=0,\ldots,t_m = m\cdot\frac{1}{M},\ldots,t_M=1$ by the procedure described in Section~\ref{sec:als}. In the following, we discuss the critical question of how to choose the samples defining the empirical loss \eqref{eq:emp_error} at any given time.

\subsection{Stochastic steps}
There are two main obstacles when solving the PDE \eqref{eq:emp_error}.
First, it is unclear how to choose the samples on which to solve the PDE. We could use uniform samples in space. 
However, this scales unfavorably.
Alternatively, we can follow \cite{balint}, which always draws from the current trained flow. This is similar to a \emph{mode-seeking} reverse KL in the context of normalizing flows \cite{papamakarios2021normalizing} and our method of choice.
Pushing samples along the trajectories of the learned vector field is later denoted by $\mathtt{ODEsolve}$.
Second, the path of $f_t$ is well-known to exhibit large jumps in the velocity field close to the final time, where relative mode weights get reassigned. 
This is called ``teleportation issue" in \cite{chemseddine2024neuralsamplingboltzmanndensities}, where it is investigated analytically. See also \cite{balint} for visualizations of this issue. 

Both obstacles can be mitigated by adding stochastic steps. This has been crucial in many sampling related works, such as \cite{arbel2021annealed,hertrich2024importancecorrectedneuraljko,wu2020stochastic}. After the TT steps, two kinds of stochastic steps are added as outlined in Algorithm~\ref{alg:train}:

\begin{itemize}
    \item \emph{Langevin steps}: Given an energy $f$ of some density $p$, we can sample asymptotically from $p$ by following the dynamics
    $$
        X_{\ell+1} = X_\ell - h\ \nabla f(X_\ell) + \sqrt{2h}\mathcal{N}(0,I),
    $$
    where $(X_\ell)_\ell$ denotes a Markov chain that converges asymptotically to samples from $p \sim e^{-f}$ ~\cite{Dalalyan2017UserfriendlyGF,miranda2024approximatinglangevinmontecarlo,zhang2023nonasymptotic}. Required for this are growth and regularity conditions on $f$ such as strong convexity and Lipschitz gradient of the energy. Only a few steps of this update rule are applied and denoted by $\mathtt{LangevinStep}$ in Algorithm~\ref{alg:train}.
    \item \emph{Resampling steps}: Since Langevin sampling struggles with mode mixing \cite{wu2020stochastic}, we also perform so-called ``resampling steps'' as commonly used in Sequential Monte Carlo methods~\cite{arbel2021annealed,del2006sequential,doucet2000sequential}.
    Here, we are given some points $\{x^{(\ell)}\}_{\ell=1}^N$ drawn from some density $q$, which can be evaluated up to the normalizing constant. Assume we target density $p$.
    Then, for each of the points $x^{(\ell)}$ the importance weights \cite{neal2001annealed} $w^{(\ell)} = \frac{p(x^{(\ell)})}{q(x^{(\ell)})}$ are calculated. Then, samples are drawn from the categorical distribution proportional to $\sum_\ell w^{(\ell)} x^{(\ell)}$.
    This however yields the ``same'' sample multiple times, which is why we also add Langevin steps after resampling steps. This procedure is denoted by $\mathtt{ResamplingStep}$. 
\end{itemize}
While the resampling steps help to overcome teleportation issues, the Langevin dynamics deter a collapse of the resampling steps to point masses. For the resampling step, we still have to discuss how $q$ should be chosen. 
Advancing step $k \rightarrow k+1$ with a known transport $T_k = \phi(t_{k+1}, \cdot) \circ \phi(t_{k}, \cdot)^{-1}$ with $(T_k)_{\#}p_k = q \approx p_{k+1}$ as in~\eqref{eq:dynamics}, the density $q$ can be written as $q(x) = p_k(T_k^{-1}(x))|\mathrm{det} \nabla T_k^{-1}(x))|$ by leveraging the change of variables formula.
The map $T_k$ given a velocity field $v$ can be defined by $T_k(x)  = x + \int_{t_k}^{t_{k+1}} v_s(x_s) \, \text{d}s$, which we approximate using our discrete formulation (and similarly for $T_k^{-1}$). 
In practice we find it beneficial to use Langevin steps before and after resampling steps.
However, assessing the effect of the Langevin layers on the density $q$ is left for future work and we only use the flow layers. 

\subsection{Adaptive TT ranks}
The training on one set of prior samples, which is transported by the vector field to serve as training samples for the TT in the next time step, is repeated to train the TTs on more samples. In each iteration, the distribution of the samples in later time steps should be closer to the interpolated distribution, since the TTs approximate the solution of the PDE increasingly better in each iteration. They therefore serve as better training samples.
When training on a new set of samples, the ALS updates each core without consideration of the information from the last iteration. To not deviate too far from the previously learned TT in the last iteration, the old and newly learned TTs are added in a weighted manner similar to an \emph{exponential moving average (EMA)}.
The TT representing the addition can exhibit doubled ranks, which can be recompressed by applying a singular value decomposition on each core and truncating small singular values (TT-SVD, \cite{O11}).
This procedure is denoted by $\mathtt{UpdateTT}$ in Algorithm~\ref{alg:train} and bears similarities with the CRAFT framework \cite{CRAFT2022}. By this, the ranks of the TT do not have to be set a priori but are adjusted dynamically during training. The resulting algorithm is described in Algorithm~\ref{alg:train}.

\begin{algorithm}
    \caption{Training the vector field}\label{alg:train}
    \textbf{Input:}  latent density $p_0$, target energy $f_1$\\
    \While{not converged}{
    sample prior $x \leftarrow \{x_{i}\}_{i\in [N]}\sim p_0$ 
    \Comment{initialize training data set}\\
    \For{$k=1,\dots,M$}{
    calculate $\partial_t f_{t_k}(x), \nabla f_t(x)$\\
    train $v_{t_k}^{\text{new}}$ with ALS to minimize \eqref{eq:quad_program} \Comment{train vector field}\\
    $v_{t_k} \leftarrow \mathtt{UpdateTT}(v_{t_k}, v_{t_k}^{\text{new}})$ \Comment{update and adjust TT ranks}\\
    $x \leftarrow \mathtt{ODEsolve}(x, t_k, t_{k+1}, v_{t_k})$\Comment{update samples for next time step}\\ 
    $x \leftarrow \mathtt{ResamplingStep}(x)$\\
    $x \leftarrow \mathtt{LangevinStep}(x)$
    }
    }
    \textbf{return:} $v_{t_1},\dots,v_{t_M}$
\end{algorithm}

%------------------------------------------------------
\section{Numerical experiments}
The proposed architecture is tested with different target distributions. For comparison with other works, the energy distance as in~\cite{sejdinovic,szekely} is used.

\paragraph{Target Distributions.}
As targets, we choose the following three distributions: 
\begin{enumerate}
    \item\label{2dgmm} \emph{Gaussian mixture with two modes}: The two-dimensional Gaussian mixture is considered with two modes. Both have same weights and variance $0.01$. One Gaussian is with mean $(2,2)$ and other with mean $(-2,-2)$.
    %defined by $$f_1 = \log\left(\exp\left(-\frac12\frac{\norm{x - (2, 2)}^2}{0.01}\right) + \exp\left(- \frac12\frac{\norm{x - (-2, -2)}^2}{0.01} \right)\right)$$.
    \item\label{2dmultimodal} \emph{Gaussian mixture with $40$ modes}: The two-dimensional Gaussian mixture as in~\cite{chemseddine2024neuralsamplingboltzmanndensities,midgley2023flowannealedimportancesampling} is considered, where $40$ different normal distributions with means uniformly distributed in $[-40,40]^2$ are mixed with equal weights. 
    %The latent distribution is chosen to be a mean zero normal distribution and variance $300$ as the initial distribution.
    \item\label{manywell} \emph{Many well problem (MWP)}: The target density in the $d$-dimensional MWP as in~\cite{chemseddine2024neuralsamplingboltzmanndensities,midgley2023flowannealedimportancesampling} consists of the multiplication of $m$ copies of the two-dimensional double well potential and $d-2m$ Gaussians given by\\
    \phantom{\qquad} $f_1^{m,d} = \sum_{i=0}^m \left(x_{2i}^4 - 6x_{2i}^2 -\frac{1}{2} x_{2i} + \frac{1}{2} x_{2i+1}^2\right) + \sum_{i=2m+1}^{d} \frac{1}{2} x_i^2 .$
\end{enumerate}

\paragraph{Methods.}
To assess the importance of the different algorithmic steps, we test the following methods:
\begin{itemize}
    \item \textbf{flow}: a pure TT normalizing flow is trained without stochastic steps.
    \item \textbf{flow+}: after the full TT flow, resampling and Langevin steps are carried out.
    \item \textbf{flow+stochastic}: a flow is trained with stochastic steps in the training process and the evaluation process after every time step as in Algorithm \ref{alg:train}.
    \item \textbf{stochastic}: the results are compared with samples generated by the same amount of stochastic steps without steps of a trained flow, where we take the ratio for the resampling step to be $\frac{p_{k+1}}{p_k}$. 
\end{itemize}
 In all  examples, the ground truth is known. The models are tuned by comparison with a validation set of samples. The tuning procedures without ground truth samples is left for future work. 

In all experiments, the basis is chosen to be an $H^2$ orthonormalized Fourier basis and
%{\color{red} ist nat etwas geschummelt, da wir das nur auf einem Intervall und nicht in R haben, kommt dann noch - das Rote auskommentieren, aber im Hinterkopf behalten}, 
the number of time steps is $M\leq35$. The number of Fourier basis coefficients is $n\leq 10$ in all experiments except for Problem 2, in which we use $n=41$. In Problems~\ref{2dgmm} and \ref{manywell}, the basis is orthonormalized on $[-5,5]^d$ and the latent distribution is chosen to be a normal distribution.
In Problem~\ref{2dmultimodal}, the parameters are $[-50,50]^2$ and we start in a zero mean Gaussian with standard deviation $500$.
%An exemplary velocity field of the trained flow can be seen in Fig.~\ref{fig:v_flow}.

\begin{figure}
    \centering
    % First row of subfigures
    \begin{subfigure}[b]{0.45\textwidth}
        \centering
        \includegraphics[width=\textwidth]{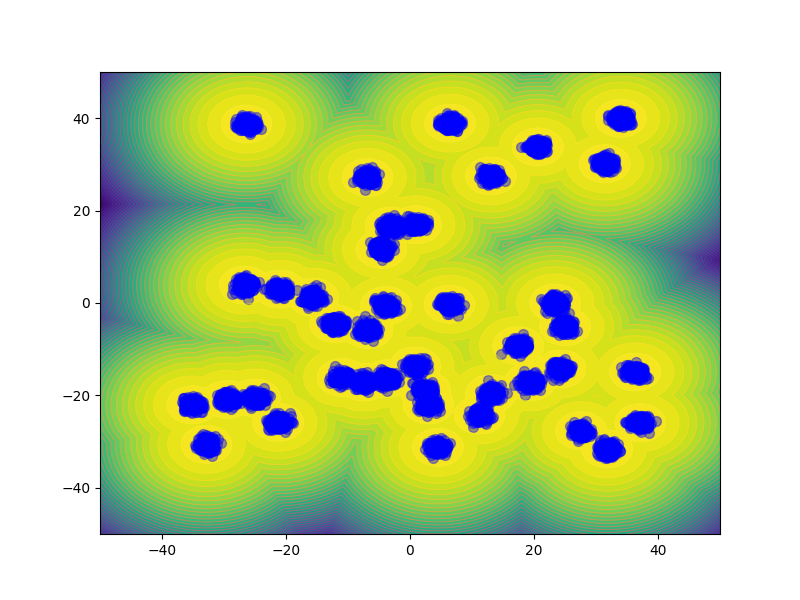} % Replace with your image
        \caption{ground truth}
        \label{fig:sub1}
    \end{subfigure}
    \hfill
    \begin{subfigure}[b]{0.45\textwidth}
        \centering
        \includegraphics[width=\textwidth]{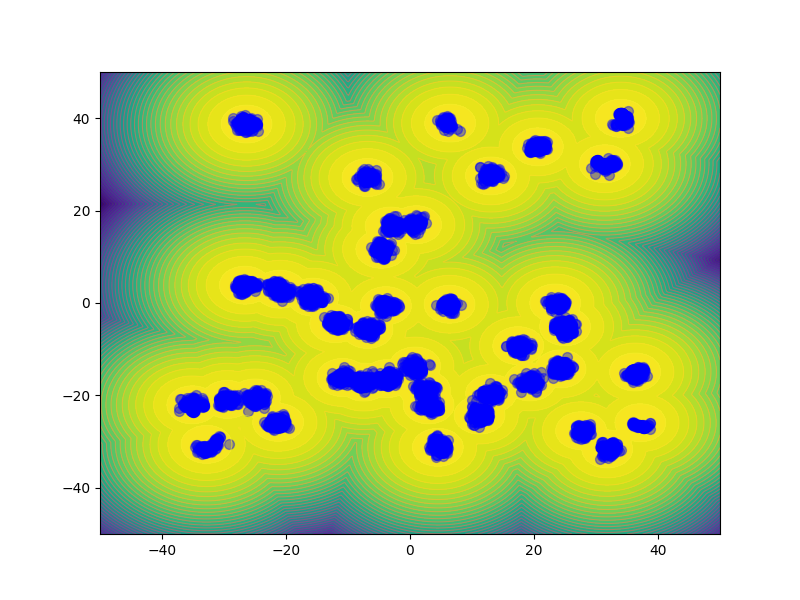} % Replace with your image
        \caption{flow+}
        \label{fig:sub2}
    \end{subfigure}
    
    % Second row of subfigures
    \begin{subfigure}[b]{0.45\textwidth}
        \centering
        \includegraphics[width=\textwidth]{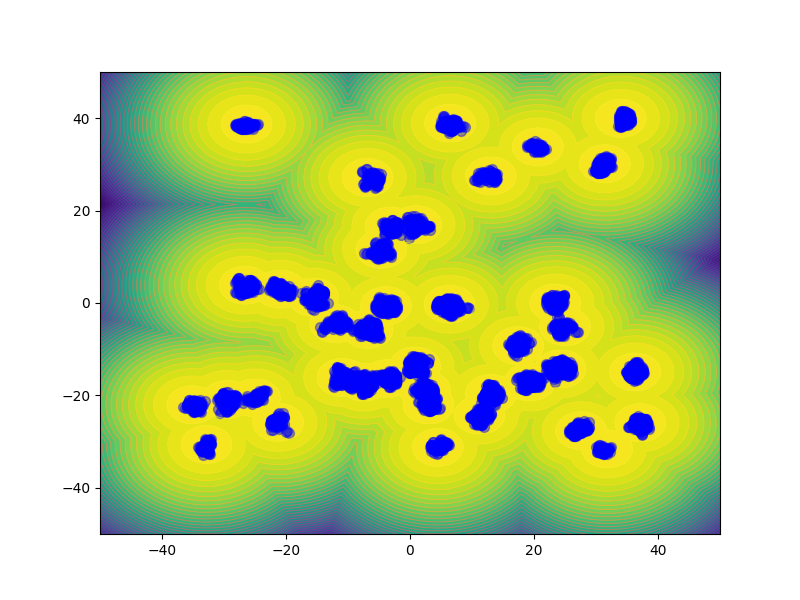} % Replace with your image
        \caption{flow+stochastics}
        \label{fig:sub3}
    \end{subfigure}
    \hfill
    \begin{subfigure}[b]{0.45\textwidth}
        \centering
        \includegraphics[width=\textwidth]{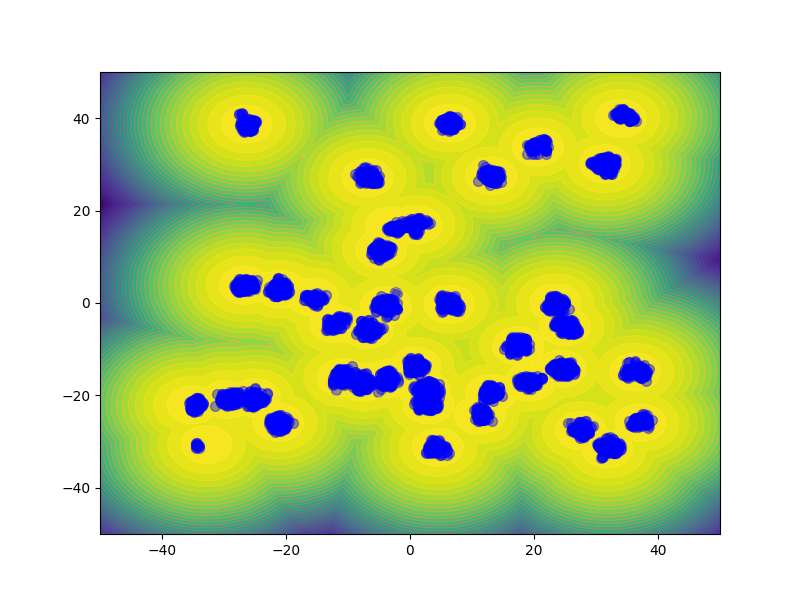} % Replace with your image
        \caption{stochastics}
        \label{fig:sub4}
    \end{subfigure}
    
    \caption{GM density sampled with the different methods. While the methods perform similarly on this problem,  the TT flow in (b) and (c) helps to distribute the mass correctly compared to a purely stochastic approach (as can e.g. be seen for the mode on the lower left part of the domain).}
    \label{fig:2x2}
\end{figure}

\begin{figure}
    \centering
     \begin{subfigure}[b]{0.49\textwidth}    \includegraphics[width=\linewidth]{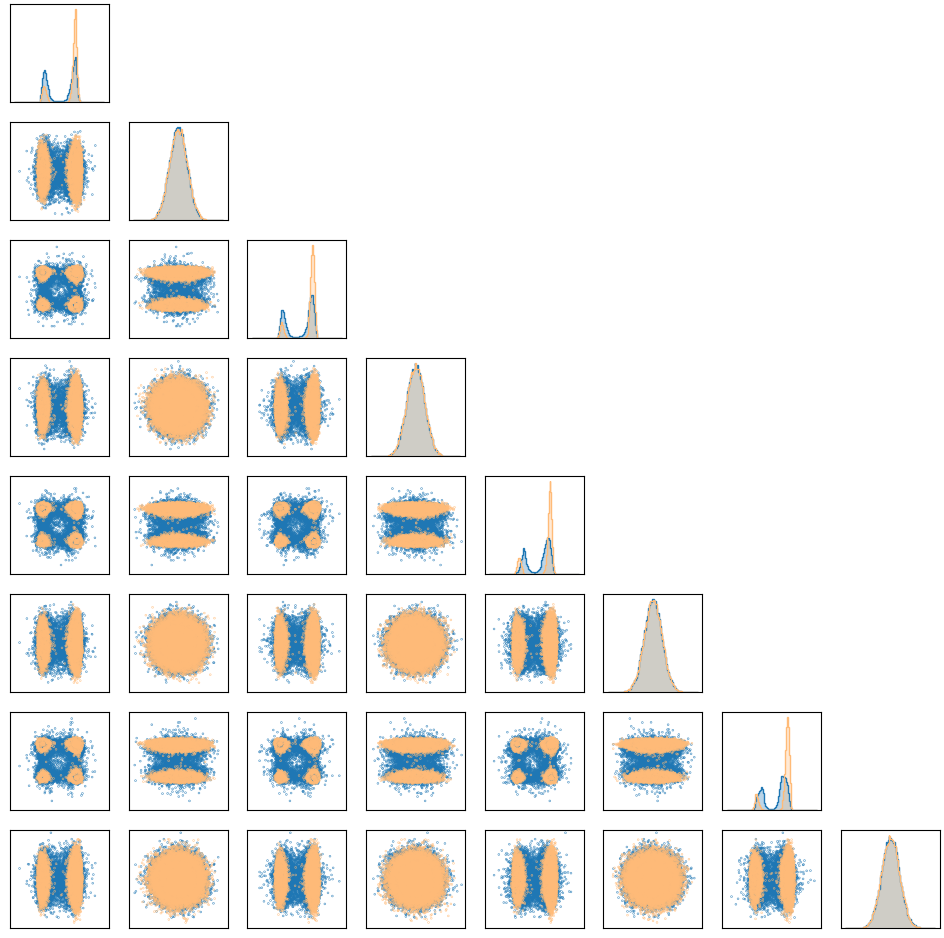}    
     \caption{flow}
    \end{subfigure}
    \begin{subfigure}[b]{0.49\textwidth}   
    \includegraphics[width=\linewidth]{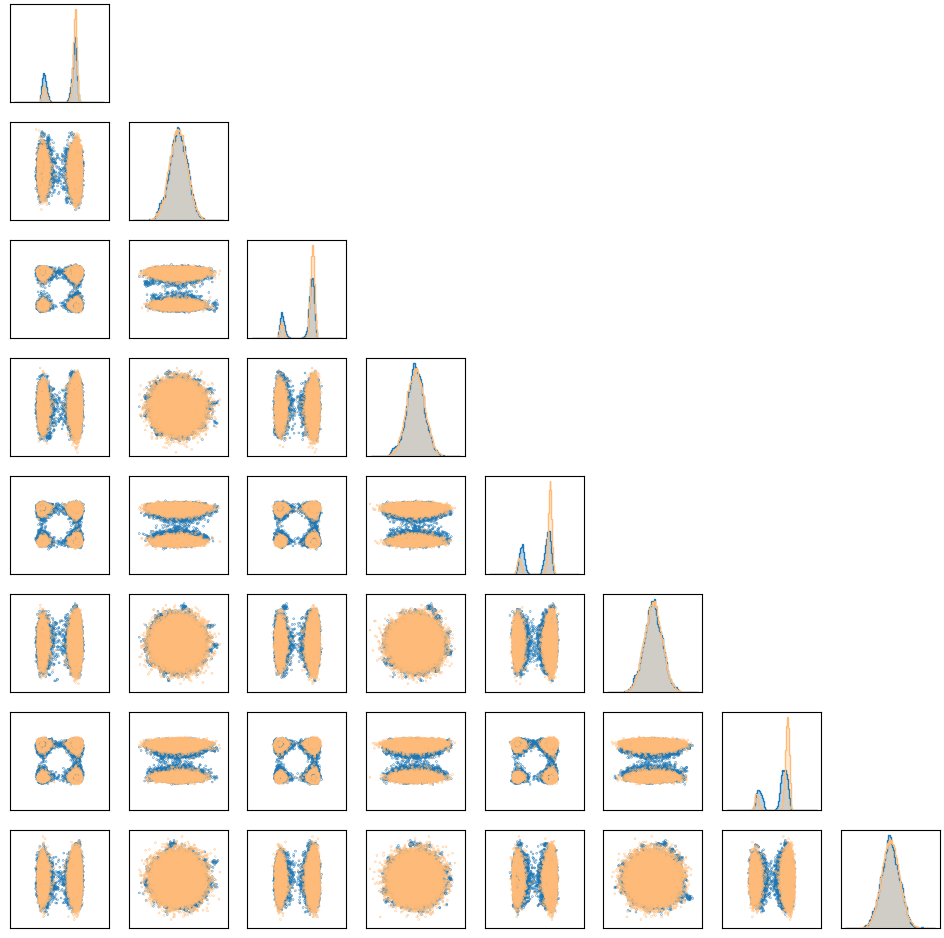}    
     \caption{flow+}
    \end{subfigure}
     \begin{subfigure}[b]{0.49\textwidth}   
    \includegraphics[width=\linewidth]{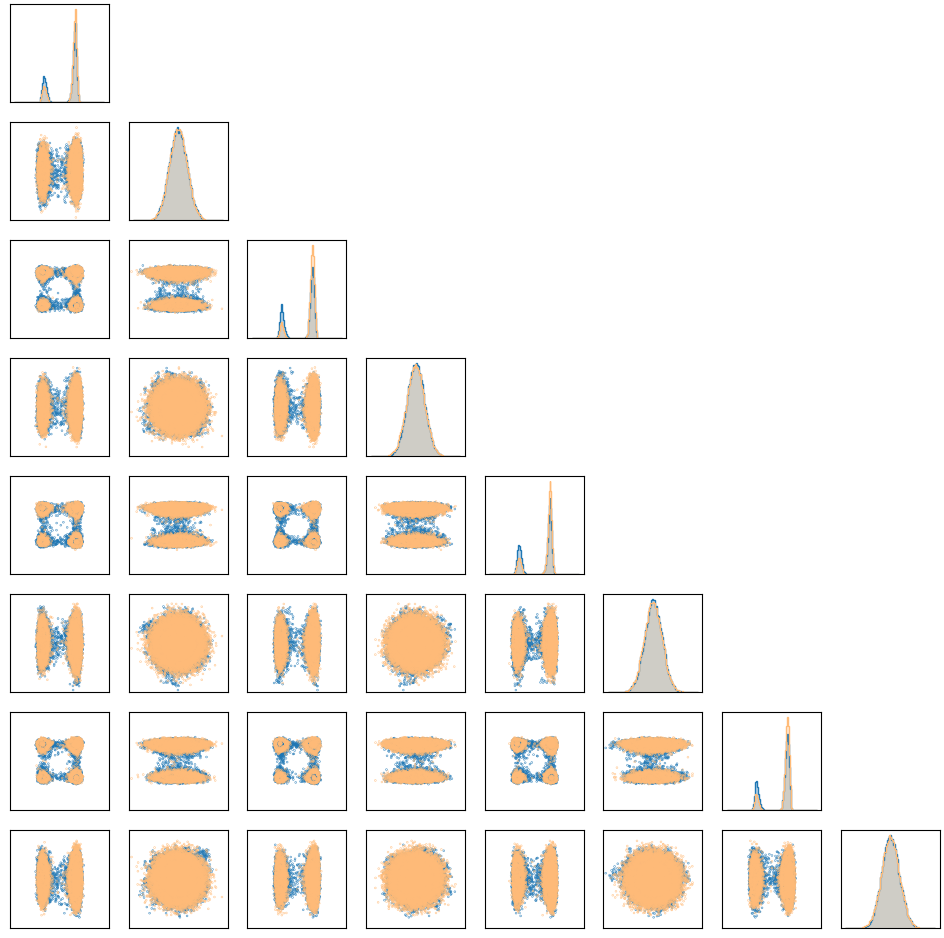} 
     \caption{stochastic}
        \end{subfigure}
     \begin{subfigure}[b]{0.49\textwidth}   
    \includegraphics[width=\linewidth]{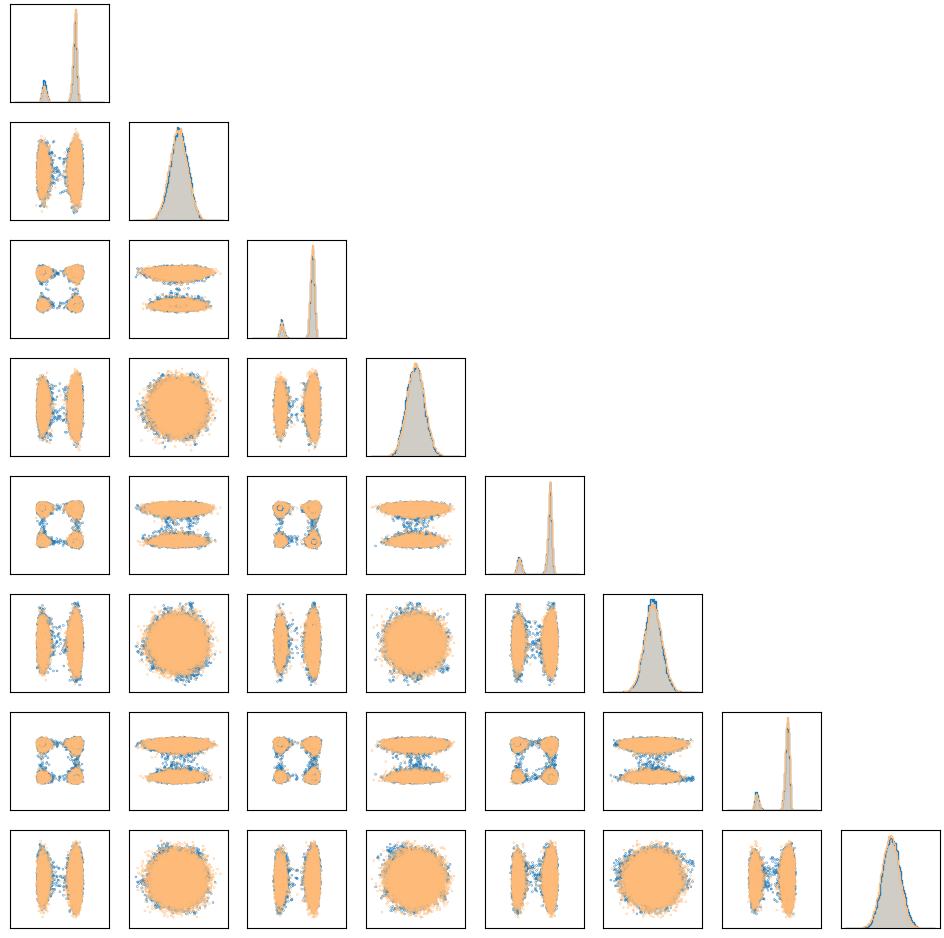} 
     \caption{flow+stochastic}
     \end{subfigure}

    \caption{Histogram corner plots showing the 1-d marginals on the diagonal and the 2-d joint distributions on the off-diagonal for the MWP $f^{4,16}_1$ for different methods (blue: competing method, orange ground truth).}
    \label{fig:many_well}
\end{figure}

Table~\ref{tab:2DgmmDifferentFlows} contains the results for the four methods and four different target distributions. 
It can be seen that the pure flow already achieves a good fit.
Especially in higher dimensions, the stochastic steps are needed to achieve very good performance.
Usually either \emph{flow+} or \emph{flow+stochastic} take the lead, where \emph{flow+stochastic} seems to perform especially well in the high-dimensional MWP examples. We also see that the flow improves upon the stochastic steps since \emph{flow+stochastic} usually outperforms \emph{stochastic}.
We show the marginal distributions of the 4 methods in Figure \ref{fig:many_well} for some marginals of the MWP. Here, continuous TT flow (\emph{flow}) struggles to separate the mass perfectly, which is resolved by the stochastic steps. Figure~\ref{fig:2x2} depicts the samples generated for the $40$-mode GM problem. 

\begin{table}
    \centering
    \caption{Average energy distance $\pm$ standard deviation over $100$ independent draws of $50.000$ samples for the methods flow, flow+, flow+stochastic and stochastic. Best method in bold.}
    \begin{tabular}{l|cccc}
        \toprule
         & flow & flow+ & flow+stochastic & stochastic\\
         \midrule
         GM 2 modes & $\ 3.7\e{-3} \pm 2\e{-4}\ $ & $\  \mathbf{1.2\textbf{e}{-3} \pm 5\textbf{e}{-4}} \ $ & $\ \mathbf{1.1\textbf{e}{-3} \pm 1\textbf{e}{-3}}\ $ & $\ 2.7\e{-3}\pm 3\e{-3}\ $ \\
         GM $40$ modes & $\ 4.5\e{-1} \pm 3\e{-2}$ & $\mathbf{\ 7.6\textbf{e}{-2} \pm 2\e{-2}}$ & $\mathbf{\ 7.6\textbf{e}{-2} \pm 4\textbf{e}{-2}}$ & $\ 1.4\e{-1} \pm 9\e{-2}$\\
         Many well $f^{4,8}_1$ & $1.6\e{-1} \pm 1\e{-4}$ & $1.2\e{-2} \pm 2\e{-4}$ & $\mathbf{1.9\textbf{e}{-3}\pm 2\textbf{e}{-4}}$ & $8.1\e{-2}\pm 9\e{-3}$ \\
         MWP $f^{4,16}_1$ & $2.1\e{-1} \pm 2\e{-3}$ & $1.3\e{-1}\pm 6\e{-3}$ & $\mathbf{6.6\e{-3} \pm 2\e{-3}}$ & $7.4\e{-2}\pm 1\e{-2}$\\
         \bottomrule
    \end{tabular}
    \label{tab:2DgmmDifferentFlows}
\end{table}

\section{Conclusions}
In this work, the continuity equation of the annealing path is solved in a functional low-rank TT format to draw samples from a given target density.
%The usual one dimensional TT architecture is extended to higher dimensions by setting the first core to a three dimensional tensor. 
In contrast to other works, a Fourier basis orthogonalized with respect to the $H^2$ scalar product is used for the FTT architecture. This choice works well to solve the PDE, while common basis functions such as polynomials perform poorly. This observation could be attributed to the lack of Lipschitz continuity of higher-order polynomials and will be examined in future work.
The TT architecture is adapted during training similar to an EMA, leading to a more stable training procedure and the advantage that the architecture does not have to be set a priori.
In future work, non-sample-based techniques could be developed to choose the best architecture during training, e.g., based on the residual of the PDE.

\section{Acknowledgements}
P.H., M.E., G.S. \& J.S. gratefully acknowledge the funding from DFG SPP 2298 "Theoretical Foundations of Deep Learning". M.E. \& D.S. acknowledge support by the ANR-DFG project \textit{COFNET: Compositional functions networks - adaptive learning for high-dimensional approximation and uncertainty quantification}. This study does not have any conflicts to disclose.
% \newpage
\bibliographystyle{abbrv}
\bibliography{lib}

\begin{thebibliography}{10}

\bibitem{Albergo_2019}
M.~Albergo, G.~Kanwar, and P.~Shanahan.
\newblock Flow-based generative models for {M}arkov chain {Monte Carlo} in lattice field theory.
\newblock {\em Physical Review D}, 100(3), Aug. 2019.

\bibitem{ambrosio2008gradient}
L.~Ambrosio, N.~Gigli, and G.~Savar{\'e}.
\newblock {\em Gradient flows: in metric spaces and in the space of probability measures}.
\newblock Springer Science \& Business Media, 2008.

\bibitem{arbel2021annealed}
M.~Arbel, A.~Matthews, and A.~Doucet.
\newblock Annealed flow transport {Monte Carlo}.
\newblock In {\em International Conference on Machine Learning}, pages 318--330. PMLR, 2021.

\bibitem{mcmc_barrier}
A.~Bandeira, A.~Maillard, R.~Nickl, and S.~Wang.
\newblock On free energy barriers in gaussian priors and failure of cold start {MCMC} for high-dimensional unimodal distributions.
\newblock {\em Philosophical transactions. Series A, Mathematical, physical, and engineering sciences}, 381:20220150, 03 2023.

\bibitem{berner2024an}
J.~Berner, L.~Richter, and K.~Ullrich.
\newblock An optimal control perspective on diffusion-based generative modeling.
\newblock {\em Transactions on Machine Learning Research}, 2024.

\bibitem{chemseddine2024neuralsamplingboltzmanndensities}
J.~Chemseddine, C.~Wald, R.~Duong, and G.~Steidl.
\newblock Neural sampling from {B}oltzmann densities: {Fisher-Rao} curves in the {W}asserstein geometry.
\newblock {\em arXiv preprint arXiv:2410.03282}, 2024.

\bibitem{chuang2022experience}
P.-Y. Chuang and L.~A. Barba.
\newblock Experience report of physics-informed neural networks in fluid simulations: pitfalls and frustration.
\newblock {\em arXiv preprint arXiv:2205.14249}, 2022.

\bibitem{Dalalyan2017UserfriendlyGF}
A.~S. Dalalyan and A.~G. Karagulyan.
\newblock User-friendly guarantees for the {Langevin Monte Carlo} with inaccurate gradient.
\newblock {\em ArXiv}, abs/1710.00095, 2017.

\bibitem{Del_Debbio_2021}
L.~Del~Debbio, J.~Marsh~Rossney, and M.~Wilson.
\newblock Efficient modeling of trivializing maps for lattice $\phi^4$ theory using normalizing flows: A first look at scalability.
\newblock {\em Physical Review D}, 104(9), Nov. 2021.

\bibitem{del2006sequential}
P.~Del~Moral, A.~Doucet, and A.~Jasra.
\newblock Sequential {Monte Carlo} samplers.
\newblock {\em Journal of the Royal Statistical Society Series B: Statist. Method.}, 68(3):411--436, 2006.

\bibitem{dolgov2021tensor}
S.~Dolgov, D.~Kalise, and K.~K. Kunisch.
\newblock Tensor decomposition methods for high-dimensional {Hamilton--Jacobi--Bellman} equations.
\newblock {\em SIAM Journal on Scientific Computing}, 43(3):A1625--A1650, 2021.

\bibitem{doucet2000sequential}
A.~Doucet, S.~Godsill, and C.~Andrieu.
\newblock On sequential monte carlo sampling methods for bayesian filtering.
\newblock {\em Statistics and Computing}, 10:197--208, 2000.

\bibitem{EigelNeumannSchneiderWolf+2019+39+53}
M.~Eigel, J.~Neumann, R.~Schneider, and S.~Wolf.
\newblock Non-intrusive tensor reconstruction for high-dimensional random pdes.
\newblock {\em Computational Methods in Applied Mathematics}, 19(1):39--53, 2019.

\bibitem{eigel2023dynamical}
M.~Eigel, R.~Schneider, and D.~Sommer.
\newblock Dynamical low-rank approximations of solutions to the {Hamilton--Jacobi--Bellman} equation.
\newblock {\em Numerical Linear Algebra with Applications}, 30(3):e2463, 2023.

\bibitem{HHS2022}
P.~Hagemann, J.~Hertrich, and G.~Steidl.
\newblock Stochastic normalizing flows for inverse problems: A {M}arkov chains viewpoint.
\newblock {\em SIAM/ASA Journal on Uncertainty Quantification}, 10(3):1162--1190, 2022.

\bibitem{hertrich2024importancecorrectedneuraljko}
J.~Hertrich and R.~Gruhlke.
\newblock Importance corrected neural {JKO} sampling.
\newblock {\em arXiv preprint arXiv:2407.20444}, 2024.

\bibitem{holtz2012alternating}
S.~Holtz, T.~Rohwedder, and R.~Schneider.
\newblock The alternating linear scheme for tensor optimization in the tensor train format.
\newblock {\em SIAM Journal on Scientific Computing}, 34(2):A683--A713, 2012.

\bibitem{holtz2012manifolds}
S.~Holtz, T.~Rohwedder, and R.~Schneider.
\newblock On manifolds of tensors of fixed {TT}-rank.
\newblock {\em Numerische Mathematik}, 120(4):701--731, 2012.

\bibitem{balint}
B.~M{\'a}t{\'e} and F.~Fleuret.
\newblock Learning interpolations between boltzmann densities.
\newblock {\em Transactions on Machine Learning Research}, 2023.

\bibitem{CRAFT2022}
A.~G. D.~G. Matthews, M.~Arbel, D.~J. Rezende, and A.~Doucet.
\newblock Continual repeated annealed flow transport {Monte Carlo}.
\newblock In {\em Proceedings of the 39th Int. Conference on Machine Learning}, Proceedings of Machine Learning Research, Jul 2022.

\bibitem{midgley2023flowannealedimportancesampling}
L.~I. Midgley, V.~Stimper, G.~N.~C. Simm, B.~Schölkopf, and J.~M. Hernández-Lobato.
\newblock Flow annealed importance sampling bootstrap.
\newblock {\em arXiv preprint arXiv:2208.01893}, 2023.

\bibitem{miranda2024approximatinglangevinmontecarlo}
C.~Miranda, J.~Schütte, D.~Sommer, and M.~Eigel.
\newblock Approximating {Langevin Monte Carlo} with {ResNet}-like neural network architectures.
\newblock {\em arXiv preprint arXiv:2311.03242}, 2024.

\bibitem{neal2001annealed}
R.~M. Neal.
\newblock Annealed importance sampling.
\newblock {\em Statist. and Comp.}, 11:125--139, 2001.

\bibitem{Nouy_2017}
A.~Nouy.
\newblock {\em Chapter 4: Low-Rank Methods for High-Dimensional Approximation and Model Order Reduction}, page 171–226.
\newblock Society for Industrial and Applied Mathematics, July 2017.

\bibitem{noe_boltzmann}
F.~Noé, S.~Olsson, J.~Köhler, and H.~Wu.
\newblock Boltzmann generators: Sampling equilibrium states of many-body systems with deep learning.
\newblock {\em Science}, 365(6457), 2019.

\bibitem{O11}
I.~V. Oseledets.
\newblock Tensor-train decomposition.
\newblock {\em SIAM Journal on Scientific Computing}, 33(5):2295--2317, 2011.

\bibitem{oseledets2013constructive}
I.~V. Oseledets.
\newblock Constructive representation of functions in low-rank tensor formats.
\newblock {\em Constructive Approximation}, 37:1--18, 2013.

\bibitem{papamakarios2021normalizing}
G.~Papamakarios, E.~Nalisnick, D.~J. Rezende, S.~Mohamed, and B.~Lakshminarayanan.
\newblock Normalizing flows for probabilistic modeling and inference.
\newblock {\em Journal of Machine Learning Research}, 22(57):1--64, 2021.

\bibitem{RAISSI2019686}
M.~Raissi, P.~Perdikaris, and G.~Karniadakis.
\newblock Physics-informed neural networks: A deep learning framework for solving forward and inverse problems involving nonlinear partial differential equations.
\newblock {\em Journal of Comp. Phys.}, 378:686--707, 2019.

\bibitem{richter2021solving}
L.~Richter, L.~Sallandt, and N.~N{\"u}sken.
\newblock Solving high-dimensional parabolic {PDEs} using the tensor train format.
\newblock In {\em International Conference on Machine Learning}, pages 8998--9009. PMLR, 2021.

\bibitem{rohwedder2013local}
T.~Rohwedder and A.~Uschmajew.
\newblock On local convergence of alternating schemes for optimization of convex problems in the tensor train format.
\newblock {\em SIAM Journal on Numerical Analysis}, 51(2):1134--1162, 2013.

\bibitem{sejdinovic}
D.~Sejdinovic, B.~Sriperumbudur, A.~Gretton, and K.~Fukumizu.
\newblock {Equivalence of distance-based and RKHS-based statistics in hypothesis testing}.
\newblock {\em The Annals of Statistics}, 41(5):2263 -- 2291, 2013.

\bibitem{sommer2024TT}
D.~Sommer, R.~Gruhlke, M.~Kirstein, M.~Eigel, and C.~Schillings.
\newblock Generative modelling with tensor train approximations of {Hamilton--Jacobi--Bellman} equations.
\newblock {\em arXiv preprint arXiv:2402.15285}, 2024.

\bibitem{sun2024dynamicalmeasuretransportneural}
J.~Sun, J.~Berner, L.~Richter, M.~Zeinhofer, J.~Müller, K.~Azizzadenesheli, and A.~Anandkumar.
\newblock Dynamical measure transport and neural {PDE} solvers for sampling.
\newblock {\em arXiv preprint arXiv:2407.07873}, 2024.

\bibitem{szekely}
G.~J. Székely and M.~L. Rizzo.
\newblock Energy statistics: A class of statistics based on distances.
\newblock {\em Journal of Statistical Planning and Inference}, 143(8):1249--1272, 2013.

\bibitem{vemuri2024functionaltensordecompositionsphysicsinformed}
S.~K. Vemuri, T.~Büchner, J.~Niebling, and J.~Denzler.
\newblock Functional tensor decompositions for physics-informed neural networks.
\newblock {\em arXiv preprint arXiv:2408.13101}, 2024.

\bibitem{wu2020stochastic}
H.~Wu, J.~K{\"o}hler, and F.~No{\'e}.
\newblock Stochastic normalizing flows.
\newblock {\em Advances in Neural Information Processing Systems}, 33:5933--5944, 2020.

\bibitem{zhang2023nonasymptotic}
Y.~Zhang, {\"O}.~D. Akyildiz, T.~Damoulas, and S.~Sabanis.
\newblock Nonasymptotic estimates for stochastic gradient {L}angevin dynamics under local conditions in nonconvex optimization.
\newblock {\em Applied Mathematics \& Optimization}, 87(2):25, 2023.

\end{thebibliography}
\end{document}